# Touch, press and stroke: a soft capacitive sensor skin

## Authors

Mirza S. Sarwar[1]*, Ryusuke Ishizaki[2], Kieran Morton[1], Claire Preston[1], Tan Nguyen[1], Xu Fan[1], Bertille Dupont[1], Leanna Hogarth[1], Takahide Yoshiike[2], Shahriar Mirabbasi[1], John D.W. Madden[1]*

## Affiliations

[1]Electrical and Computer Engineering
Advanced Materials & Process Engineering Laboratory
University of British Columbia
Vancouver V6T 1Z4
Canada
Email: jmadden@ece.ubc.ca, mirzas@ece.ubc.ca

[2]Frontier Robotics
Innovative Research Excellence
Honda R&D Co., Ltd
8-1 Honcho, Wako-shi
Saitama
351-0188
Japan

## Abstract

Soft sensors that can discriminate shear and normal force could help provide machines the fine control desirable for safe and effective physical interactions with people. A capacitive sensor is made for this purpose, composed of patterned elastomer and containing both fixed and sliding pillars that allow the sensor to deform and buckle, much like skin itself. The sensor differentiates between simultaneously applied pressure and shear. In addition, finger proximity is detectable up to 15 mm, with a pressure and shear sensitivity of 1 kPa and a displacement resolution of 50 μm. The operation is demonstrated on a simple gripper holding a cup. The combination of features and the straightforward fabrication method make this sensor a candidate for implementation as a sensing skin for humanoid robotics applications.

## Summary

A 3-axis capacitive sensor with a dielectric composed of elastomer pillars creates a skin-like deformation that allows detection of approach, light touch, pressure and shear.

## MAIN TEXT

### Introduction

To accommodate for complex interactions between humans and robots, it is important to design a method for touch identification that can be active on fingertips and other sensing surfaces. Ideally, the approach will be scalable to cover over most of a robot's surface area, forming an artificial or electronic skin (*1*, *2*). Such a technology is also sought for neurally controlled prosthetic devices to enhance motor control (*3*, *4*). The functional requirements of an artificial skin include the ability to sense and differentiate tactile stimuli such as light touch, pressure and shear (*1*). Having a smooth and soft skin, rather than a hard or bumpy surface, helps make the surface more lifelike, while the compliance allows for lower bandwidth control systems.

There is a plethora of work on flexible touch and pressure sensors. Transducers are capacitive, resistive, or piezoelectric. Flexibility is incorporated using thin polymer substrates and deformable electrode materials such as silver nanowires (*5*), carbon nanotubes (*6–8*), liquid metals (*9*, *10*), and hydrogels (*11*). Silicone based elastomer substrates such as Ecoflex (*9*, *12*) and PDMS (*13*) have been used to make the substrate stretchable. In addition to normal pressure sensing capabilities, shear sensing is an important ability to possess for an artificial skin. Differentiation of combined pressure and shear stimuli into their vector components increases dexterity. Shear sensing capability enables the user to adjust normal force so that a delicate object, such as an egg, will not slip when grasped, but also is not so large that there is damage (*14*). Altun *et al.* (*15*) emphasise the benefits of shear sensing to recognize human touch gestures on a robot.

Several sensors employ a protruding feature at each sensing location. These protrusions are torqued when sheared (*13*, *16*, *17*). The resulting angular deflection provides a deformation of the underlying material, which is detected capacitively within these bumps.

Shear sensors are typically capacitive (*18*, *19*) or piezoresistive (*20*). Several studies have been done on silicon-based tactile sensors incorporating shear detection, but these tend to be too stiff or brittle to enable highly conformable sensor arrays (*21*, *22*) and require involved fabrication processes (*13*). Most soft sensors do not measure both shear and pressure. In those that do, these forces are often not discriminated (*6*)(*23*). There are several soft sensors that have successfully discriminated between forces. Charalambides *et al.* (*24*) demonstrated a solid PDMS-based shear sensor, which had relatively low sensitivity due to its geometry and elastic stiffness. In the present work we introduce a new electrode and dielectric geometry that increases sensitivity significantly in the shear direction, and by an order of magnitude in the normal direction. More recently, Boutry *et al.* (*25*) demonstrated an elegant array of bumps, each patterned with five thin capacitive electrodes. A soft, flat surface sits above the small domes, which has five perpendicular electrodes that overlap with the electrodes below. The pattern of response produced when the two layers move relative to each other enables the differentiation between simultaneous pressure and shear in this soft sensor. However, a single pressure/shear sensing location requires interpretation of data from a 5 x 5 array of capacitive sensors, which is a challenge for scalability to larger arrays, and quantitative differentiation has not yet been shown. More recently, a 2x2 array of trapezoidal PDMS protrusions were coated with PVDF film and a flexible PCB (*26*). By comparing the voltage output from the faces of each trapezoid the direction of applied force is determined. The sensor is relatively stiff, with minimum detectable shear and normal stresses of 45 kPa and 100 kPa, respectively. These sensitivities are poor compared to the sensitivity of skin itself, where displacements of 10 micrometers are detected at low frequencies (*27*), corresponding to pressures of about 1 kPa. Xela uSkin (*28*) is a commercial normal and shear force sensor that can be expanded into an array of up to 4 x 6, and is sensitive to applied forces in the right range - down to approximately 10 mN or 0.45 kPa in normal force. It has a lateral spatial resolution of 4.7 mm per taxel. It is a magnetic sensor composed of a rubber coating mounted on a rigid chip. The surface is relatively soft, though stiffer than skin. The underlying chip (one per taxel) adds to cost and limits bending, limiting the curvature of the surface on which it can be mounted. Unlike skin, it is not readily stretchable, and, while it can bend, each taxel is stiff. These properties make it less tractable for use on curved surfaces such as fingertips.

We present a soft and stretchable sensor that is unique in its ability to sense proximity, pressure and shear. It allows differentiation between these stimuli and provides directional information regarding the shear. Its non-linear stiffness (*29*) and ability to buckle (Figure 1A left) simulates human skin – a potential advantage given that matching constitutive behaviour is considered important in current advanced humanoid robots, such as that of Sophia (Hanson Robotics) (*30*) or iCub (EU project RobotCub) (*31*) to be more realistic human synths (*32*). Unlike most previous work, the surface of the sensor is smooth. It is sensitive to the approach of a finger, or in general any grounded conductor, but is relatively insensitive to insulating materials. The sensor uses a combination of mutual capacitance to detect proximity and a light touch (*33*) and overlap capacitance to detect pressure and shear. The shear and pressure magnitudes are output as separate numeric quantities and therefore are easy to interpret and implement in automated systems, unlike more complex sensors that require pattern analysis (*25*) and a large number of data channels for a single sensing taxel. The novelty here is threefold: 1) it is the first demonstration of a soft sensor to be able to detect simultaneous application of pressure and shear stimuli quantitatively , 2) unlike most sensors in the literature, which detect shear by torqueing a protrusion, in this research the electrodes are laterally displaced, and 3) the dielectric architecture of this sensor provides a closer mechanical analogue to skin and also enables buckling - like human skin - which helps localize shear.

**Results**

A set of five independently connected electrodes is used for the capacitive sensing, enabling displacements and forces applied to the surface of the sensor to be resolved in 3 dimensions. Four top sense electrodes (*E1~E4*) are capacitively coupled to one bottom ground electrode, as shown in Fig. 1B, forming capacitances *C1~C4*. The top and bottom electrodes are coupled by electric fields both going directly through the dielectric (X1) and projecting outwards (X2), as shown in Fig. 1C. The capacitances between each of the top electrodes and the bottom reference are sequentially measured. These change in a way that depends on the nature of the stimulus.

The working principle is as follows. A human finger in proximity or lightly touching acts as a virtual ground and decouples the projected electric fields (labeled X3 in Fig. 1C), thereby reducing all the capacitances *C1~C4* in a similar manner, as previously demonstrated (*33*) and shown in Fig. 2B. Upon application of a pressure, all four top active electrodes, *E1~E4*, are displaced towards the bottom electrode, thereby increasing all four capacitances, *C1~C4* (Fig. 1D), as demonstrated in Fig. 2D. Applying a shear force in the positive x-axis moves *E3* to the right, increasing the overlapping area with the bottom electrode, and in turn increasing *C3*. *E1* also moves the to the right, decreasing overlap area with bottom electrode, in turn decreasing *C1* (Fig. 1E). The shear in this axis has minimal effects on *C2* and *C4* as their overlap with the bottom electrode does not change, as shown in Fig. 2E. This combination of changes in *C1* to *C4* is characteristic of a shear in the positive x-axis and provides information on the magnitude of the shear and its direction.

Mathematically we can separate the effects of shear and normal force. When normal force is applied, all capacitances increase, while the shear effects are proportional to the difference in capacitances. For example, assuming a parallel plate model, the shear displacement magnitude in the axis parallel to the $C_1$, $C_3$ axis is proportional to the expression $(C_3 C_1' - C_1 C_3')/(C_1' + C_3')$, where $C_1'$ and $C_3'$ are the values after the application of the shear. This equation is valid even with the application of a normal force. The derivation is presented in the supplementary document. The shear magnitude in the perpendicular direction is given by replacing $C_1$ and $C_3$ with their counterparts $C_2$ and $C_4$. The normal displacement of the dielectric is proportional to $(\Delta C_1 + \Delta C_2 + \Delta C_3 + \Delta C_4)/(C_1' + C_2' + C_3' + C_4')$, where $\Delta C_i$ represents the change in capacitance, $C_i' - C_i$. This is valid even with a simultaneous shear applied. Table 1 summarizes the behavior of the capacitances in response to proximity/light touch, normal force and two axes of shear. The overall change in capacitance of each electrode is the superposition of these individual changes.

To obtain a larger change in capacitance, as well as localize the effect of shear and simulate the buckling and stretching of skin, a novel dielectric architecture is used, as shown in Fig. 1F. The dielectric consists of elastomer pillars supporting the top layer, and surrounded by air. There are two types of pillars – square ones, with the electrodes *E1~E4* located on top of them (shown in blue), and X-shaped pillars (shown in pink). The spacing and aspect ratio of the square pillars are such that they can easily bend upon application of a shear at the top surface, making the device more sensitive compared to when a solid layer of elastomer is used. The square pillars are anchored on both the top and bottom layers, while the X pillars are only anchored to the bottom layer, as shown in Fig. 1G. This allows for a smooth top layer by preventing it from collapsing in regions not supported by square pillars, shown in Fig. 1G, while simultaneously enabling a sliding motion of the top layer upon application of a shear. It is this feature that gives the sensor the ability to buckle and stretch in a localized region, mimicking human skin, as shown in Fig. 1A (right).

The materials and methods are deliberately chosen to be low cost and appropriate for large-format mass production. A simple three-step process mold-pattern-bond (MPB **Fig. 3**A to D) produces a unibody sensor. The elastomer used is Ecoflex™ 00-30. It is commonly found in toys, face masks, and costumes, providing a skin like feel. The stretchable electrodes are carbon black mixed with Ecoflex™. Although the resistivity of the carbon black based stretchable conductor is large (approximately 0.1 Ω-m), it is sufficient for capacitive sensing since this requires only very small currents.

In the MPB process developed to fabricate the sensor, Ecoflex™ 00-30 is first cured in the top and bottom molds (**Fig. 3**A) to build the top and bottom segments, with the two types of pillars in the two molds being 1.5 mm tall, beneath a 0.3 mm thick top layer. The air pockets are 1.5 mm tall, the square pillars are 3 mm x 3 mm, and the X-pillars are 5.8 mm long on each leg. The electrodes are then patterned with carbon black containing uncured Ecoflex™ using shadow masks made from cut transparency film (Staples multipurpose transparency film, 120 μm thick) shown in **Fig. 3**B. Once patterned, the electrodes (top 4 being 3 mm x 3 mm and bottom one 9 mm x 9 mm) are covered with an encapsulating layer of Ecoflex™ (**Fig. 3**C) approximately 300 μm thick. Half of the area of each of the top 4 electrodes overlaps with the bottom electrode midway along the four edges (Fig. 1B). The top segment is then bonded to the bottom by applying a thin layer of uncured Ecoflex™ on the bottom of the square pillars only (**Fig. 3**D). The sensor is very soft, with an effective elastic modulus for compression of approximately 160 kPa (skin has a modulus of ~ 400 kPa (*34*)). A stress-strain plot for our sensor is shown in Fig. 2A, demonstrating a low effective elastic modulus *E1* at small strains and a higher modulus *E2* for larger strains, similar to human skin (*29*).

Fig. 2B is a demonstration of sensor response to the approach and contact of a finger. The four bars in the map correspond to the four capacitances *C1~C4*. Upon approach, the projected field couples increasingly with the finger and consequently the coupling between the excitation and sense electrodes will decrease. This causes all four capacitances *C1~C4* to decrease (Fig. 2A). The maps in Fig. 2B and Fig. 2C show the % decrease in the positive y-axis. The capacitances reach a minimum with a 10-12% decrease upon light contact, as shown in Fig. 2C. The human finger acts as a virtual ground, which leads to the decrease in capacitance. The same effect is observed on the approach of a grounded metal. When approached or touched by a plastic object with a size similar to a finger, the capacitance shows a modest increase (up to 3%). The response is small and in the opposite in direction, as capacitance is increased due to the higher effective dielectric constant. The sensitivity to human approach may be valuable in robotics applications where the robot is expected to interact with humans in a delicate manner. Sensing can be done through clothing and is sensitive at a distance of 1.5 cm for electrodes of this scale. This ability to sense proximity could also be valuable in control, providing advance warning of a collision, or helping guide grasp.

Upon application of force normal to the sensor surface, the dielectric thickness is decreased, and all four capacitances increase, as shown in Fig. 2D. Fig. 2E shows the application of a shear and the sensor's response to it. The applied shear buckles the skin on the leading edge and stretches the skin on the trailing edge. This is achieved by the dielectric architecture in Fig. 1F, which enables sliding of non-bonded surfaces, and separation when under compression - much like skin itself. The capacitance response to a shear is shown in the map in Fig. 2E. The capacitance at the trailing edge of the shear increases due to the increase in overlap area, as illustrated in Fig. 1E. Also seen in Figure 2E is the decrease in capacitance at the leading edge of the sheared sensor as overlap with the reference is reduced. The two responses are not perfectly equal and opposite, due to the superimposed normal force that is applied during the act of shearing. The capacitances perpendicular to the axis of the shear force increase slightly due to this downward deformation. In this way the combined changes of the four capacitances provides information about both pressure and shear, including the direction of shear.

One key ability of this sensor architecture is to simultaneously detect and measure both the normal force and horizontal forces. This is done using the expressions:

$$normal\ strain = \frac{\eta}{d} = \frac{\Delta C_1 + \Delta C_2 + \Delta C_3 + \Delta C_4}{C_1' + C_2' + C_3' + C_4'} \quad (1),$$

$$shear\ strain\ along\ C_1 <-> C_3 = \lambda = \frac{d}{\epsilon W} \frac{C_3 C_1' - C_1 C_3'}{C_3' + C_1'} \quad (2),\ and$$

$$shear\ strain\ along\ C_2 <-> C_4 = \lambda = \frac{d}{\epsilon W} \frac{C_4 C_2' - C_2 C_4'}{C_4' + C_2'} \quad . \quad (3)$$

The derivations of the equations are shown in the supplementary. Here, $C_1$, $C_2$, $C_3$ and $C_4$ are the baseline capacitance values of the four capacitors, and $C_1'$, $C_2'$, $C_3'$ and $C_4'$ are the new capacitance values after applying a stimulus. Equations (2) and (3) provide the shear displacement along the $C_1 \sim C_3$ and the $C_2 \sim C_4$ axes. This set of three equations predicts the shear displacement along the two horizontal axes, and the normal strain, with normal force and shear being largely independent of each other. In order to estimate applied forces from capacitance, the calibration curves shown in Fig. 4 are used, as is now explained.

Like commercial force sensors, including the ATI Nano 17 multi-axis force sensor used here for calibration, the new soft sensor is able differentiate between simultaneously applied shear and normal forces. To demonstrate this, the shear and normal force results obtained from the ATI Nano 17 are compared to those predicted by our capacitive sensor. The experimental setup is shown in Fig. 4A. Normal force and % change in capacitance ($\Delta C/C_0$) for the four capacitive electrodes are plotted in Fig. 4B. We see change in capacitance increase monotonically with increasing applied force for all four electrodes, as is expected given that the separation between capacitor plates is decreasing. The slopes of the responses flatten with increasing force, consistent with the increasing stiffness of the sensor with increasing force (Fig. 2A). There is significant variation between electrodes in the $\Delta C/C_0$ magnitude and slope, shown in Fig. 4B. When looking instead at raw change in capacitance, the changes are much more consistent, Fig. S2. A major contributor to the variation in Fig. 4B is parasitic coupling, which is different for each capacitor due to the asymmetric interconnect design (discussed in the supplementary). This leads to a difference in baseline capacitance ($C1 = 44$ pF. $C2 = 49$ pF, $C3 = 49$ pF, $C4 = 38$ pF). Adding a ground shield to reduce the unwanted capacitive coupling with the interconnects, using the architecture shown in Fig. S6, the variation between electrodes is substantially reduced, Fig. 4C. The ground shielding lowers the baseline capacitances and also brings the electrode capacitance closer to each other ($C1 = 14$ pF, $C2 = 13$ pF, $C3 = 14$ pF, $C4 = 15$ pF).

Stresses of less than 1 kPa are detectable. The sensitivity is 2.8% change per kPa for low pressures. A very light press using a fingertip is approximately 50 grams, which corresponds to a pressure of about 5 kPa. The sensitivity drops with increasing load, Fig. S2-3, consistent with the increasing stiffness of the structure. At 80 kPa, it is 0.3%/kPa.

To illustrate the sensor's ability to detect normal force independent of shear forces applied, the normal strain calculated using equation (1) is plotted against the applied normal strain. The summation of the responses of the four electrode capacitances results in a single curve relating applied compressive pressure and capacitance, as shown in Fig. 4D. This curve remains essentially constant, even under simultaneous horizontal (shear) forces of varying magnitude. It is observed that all four curves, each associated with different levels of shear force, coincide with each other. The normal force detection is independent of the level of shear force applied, showing that reliable pressure readings can be obtained.

When normal strains are in excess of 30%, significant viscoelastic responses are observed, Fig. S4. However, the measurements are very stable relative to shear strain, as shown in Figure S5.

In order to characterize the shear force sensitivity, a horizontal force is applied by moving the stage in either the X or Y axes. Upon application of shear, the surface of the sensor buckles, similar to human skin, as shown in Fig. 1G. Images looking up through the bottom of the sensor in the unsheared and sheared states, are shown in Fig. 4E. The calculated shear displacement from the sensor is plotted against the applied shear displacement in Fig 4F-G for displacements along the X-axis and Y-axis. respectively. There are four sets of plots in each figure, with each set corresponding to a different normal force applied simultaneously to the shear force. The fact that all the four sets of plots coincide with each other demonstrates that the sensor is successfully able to differentiate between the normal and shear forces. The X and Y axis shear are nearly independent, with some X to Y cross-over seen in Figure 4F (10%). This is likely due to imperfect shielding, leaving some coupling between electrode E2 and E4 with the adjacent traces. Using results from the X-axis shear magnitude and Y-axis shear magnitude, the resultant shear magnitude and angle can be estimated, as demonstrated in the supplementary video, Movie S1.

The sensitivity to shear forces is illustrated in Fig. 4H and Movie S2 in the supplementary. Here the sensor is implemented on a gripper, which is used to grab a light Styrofoam cup. The sensor is mounted on a flex printed circuit board. As shown in the supplementary material (Fig. S7), the flex PCB contains the four sense electrodes (*E1-E4*) and all the readout hardware. The hardware is concealed inside the gripper, as shown in the video Movie S2. The video demonstrates the sensor's ability to detect the shear forces as the cup gets heavier due to the addition of water. This is a key offering of this sensor – its ability to detect the changes in the shear forces as the cup gets heavier; an activity requiring high multi-axis sensitivity for recognition. This feature is a step towards mass estimation and slip detection, and in turn dextrous manipulation of objects.

Minimum shear forces detected are 0.2 N (1 kPa), corresponding to shear strain of 13%, showing a sensitivity of ~0.3 % change in capacitance per change in strain (or ~20 % change in capacitance per Newton, and 5 % change/kPa), with a maximum force of 0.8 N (4.1 kPa), corresponding to a shear strain of up to 53%.

The supplementary video, Movie S1, shows the ability of the sensor to provide the magnitude of the normal displacement, the horizontal displacement magnitude and even the angle, with the simultaneous application of forces in all 3 axes. A strawberry was used to apply forces to the sensor. The strawberry, despite being relatively soft, was not physically degraded by the interactions with the sensor. This is important for applications that require a gentle interaction such as picking fruit (*35*).

Proximity to the sensor of a grounded object produces a clearly detectable signal out to a distance of 15 mm. This range is observed through the approach of a finger towards the surface of the sensor, acting as an approaching cylindrical grounded object. The response of the sensor to the approach of a finger from a maximum distance of 15 mm is illustrated in Fig. 5A, with the approaching finger shown in Fig. 5B. Due to the electric field siphoning by the approaching finger, a decrease in capacitance occurred up to a maximum average $\Delta C/C_0$ of 14.7%, measured across the four upper taxels of the sensor and using the capacitance reading at 15 mm as the baseline. This proximity is sufficient for anticipating incoming objects, and identifying light touch on the sensor surface with very low normal force.

Our sensor provides the magnitude and direction of forces in 3 axes, without need for pattern recognition (*25*).

Although shear, proximity and touch can all be detected, there is a region of ambiguity between pressure and proximity. An approaching finger or other conductor causes the capacitance to decrease. It increases upon retraction of the finger. The same response is seen if the finger approaches and pushes very lightly on the sensor, with capacitance rising after contact due to compression of the capacitors. This is illustrated in the supplementary in Fig. S8. In order to resolve this ambiguity between light pressure and proximity, a conductor plane, connected to ground, is added on top of the sensor, removing the mutual capacitive coupling between the electrodes and the approaching object or person. In order to retain the ability to detect proximity, the conductor plane is multiplexed to a self-capacitance readout IC (FDC1004). This is illustrated in Fig. S9. Movie S3 demonstrates the sensor detecting approach, light touch, and normal force, combined shear and normal force. There is no effect of proximity on force detection, and vice-versa. The configuration allows detection through clothing, and the registration of contact that is below the resolution of the normal and shear force sensors.

## Discussion

The soft capacitive sensor distinguishes normal force from shear, and provides a numerical value for both. It also enables proximity detection, and has skin-like properties. The mechanoreceptors of human skin remain more sensitive, detecting forces as small as ~ 30 mN (*36*), three quarters the weight of a sheet of paper, and corresponding to a pressure of ~300 Pa – so there is still work to do to match this sensitivity. The dielectric architecture allows plenty of room for scaling up or down, however. Changes in sensitivity can be achieved by using materials of different intrinsic stiffnesses, as well as by varying the geometry and size of the features – creating the promise of both smaller, higher resolution sensors, and bigger, higher force versions. An attractive feature of any capacitive technology is that its spatial resolution scales in proportion to the dimensions of the electrodes. On the larger scale, previous work (*33*) suggests that the active area can be hundreds of square meters, which can be low in cost due to the inexpensive materials employed.

Next steps in sensor development include coupling the sensors with control to enable the manipulation of delicate objects. There is also the potential to fabricate and test arrays of sensors to form a full skin, then connect them to a control system to enable large-area skin sensing. Effective scaling to arrays would require a reconfiguration of the sensor design described here to maximize density of sensing taxels while keeping the number of connections required manageable. One potential approach to this scaling could be arranging the taxels into grids of overlapping taxels, where vertical and horizontal traces could be shared across taxels. Success in this area would enable robots and prosthetics to interact more effectively with their environments.

## Materials and Methods

### Sensor Fabrication

A FormLabs Form2 Printer was used to print two molds (Grey V4 material). Ecoflex$^{TM}$ 00-30 (Smooth-On two-part silicone elastomer) was mixed in a 1 (part A):1 (part B) mass ratio for 3 minutes and degassed to remove air bubbles. The mixture was poured into the molds and levelled with a glass slide, degassed once more for 2 minutes and placed in an oven at 60°C for 10 minutes.

Masks for patterning the electrodes were laser cut from Staples multipurpose transparency sheets (120 μm in thickness). Each mask was manually aligned with the dielectric patterns on top of the cured Ecoflex mold layers while still in the mold, with the 4-electrode pattern laid down on the square-pillar piece and the bottom electrode laid down on the X-pillar dielectric. Carbon black (H30253 Carbon Black Super P® Conductive, Alfa Aesar) was mixed in a Thinky ARE-310 mixer with Ecoflex 00-30 in a 2(carbon black):10(part A):10(part B) mass ratio for 4 minutes, then spread over the aligned mask exposed regions to cover the electrode region. A glass slide was used to level the conductive paste and remove excess material before liftoff of the mask, leaving the patterned electrode.

Ecoflex 00-30 uncured mixture was prepared using the same mixing and degassing method as above and spin-coated at 300 RPM on top of the patterned electrodes in the molds for 60 seconds to achieve an encapsulating layer of ~400 μm thickness. The molds were again placed in the oven for 10 minutes to cure the top layer.

The two molded pieces with patterned and encapsulating electrodes were peeled off of the molds. A 1:1 Ecoflex mixture was prepared and spin-coated onto a petri dish at 300 RPM for 60 seconds. The square-pillar layer was laid gently pillar-side-down on top of this uncured layer and lifted off to form an adhesive layer on the bottom of the pillars. The dielectric piece with X-pillars was placed X side up and the square pillar piece was manually aligned such that the pillars fell in the center of the gaps between the crosses and pressed down to adhere the two layers and form the full sensor. The sensor was cured in the oven for 10 minutes at 60°C to finish the curing process.

**Sensor readout**

An Analog Devices CDC chip (AD7745) was used to read the capacitance. The CDC has a similar working principle to a Delta-Sigma analog-to-digital converter (ADC). One multiplexer (MAX 4518) was used to cycle through the four top electrodes, and the output of the multiplexer was fed to the CDC. The microcontroller (Arduino Mega 2560) controlled the multiplexer and CDC and fed the final data through a serial port into a computer where digital values of the capacitance could be acquired and displayed. For the self-capacitance augmentation an FDC1004 CDC is used in addition to the AD7745.

**Sensor Characterization**

The sensor is placed on a box containing the readout circuit, and is fixed on a 3-axis stage (Thorlabs NanoMax), as is shown in Fig. 4A. A 14 mm x 14 mm 3D printed cover is attached on to the ATI load cell, acting as the indenter plate that applied force to the sensor. Four levelling screws are used to ensure the ATI sensor is flat relative to the surface of the sensor, as any tilt will bring about unwanted torque. The soft sensor is lifted up to contact the ATI sensor using the stage. Characterization is initiated when the ATI sensor is just touching the test sensor and applying a small force (typically 0.1 N).

**Acknowledgments**

**General**: We thank Alex Abulnaga for helping with fabrication and characterization of the sensor.

**Funding:** This work was supported by an NSERC Discovery and NSERC Collaborative Research and Development grant. M.S.S. was funded by an NSERC CGS award.

## Figures and Tables

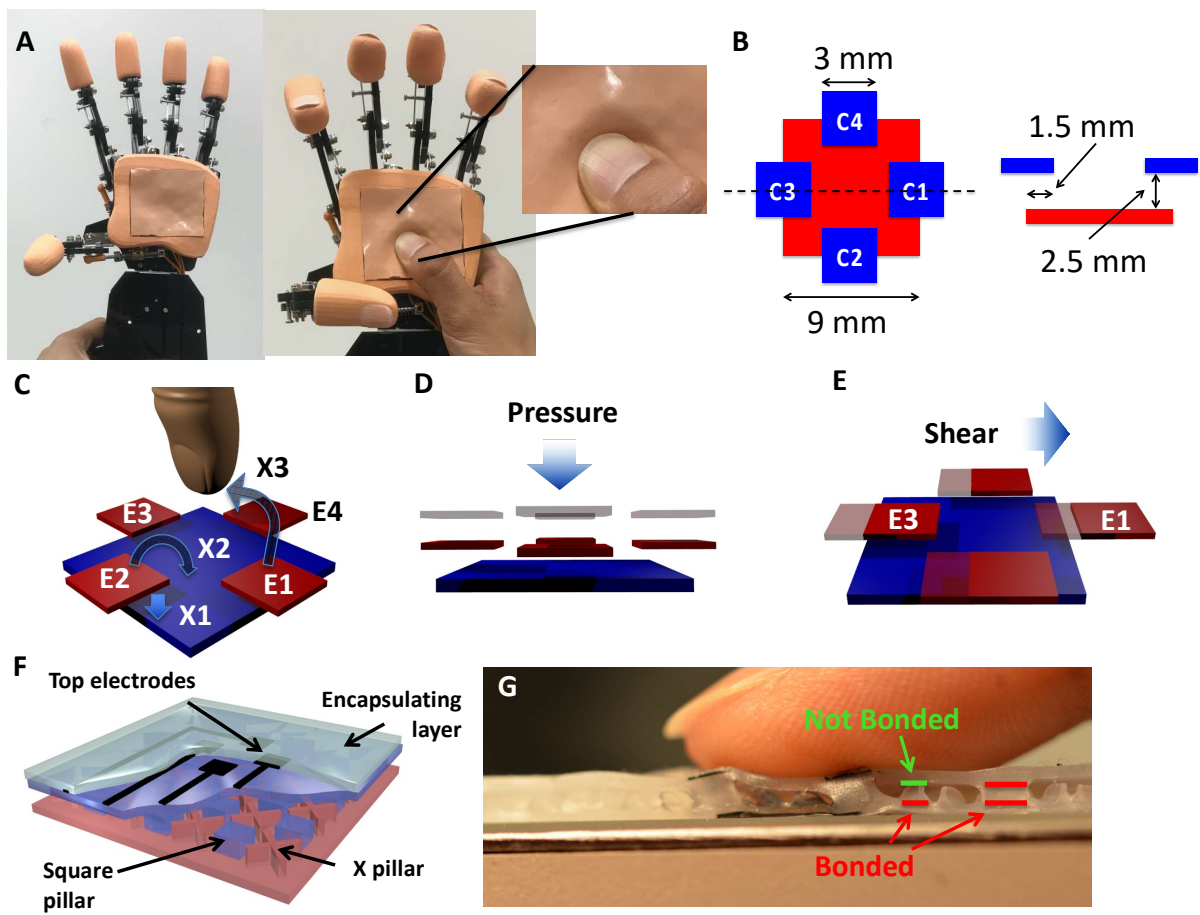

**Fig. 1. The sensor and working principle. (A)** Sensor on a robot palm (left) and sensor on robot palm sheared showing buckling similar to human palm (right) **(B)** Top view of electrode architecture (left) and cross-section (right) **(C)** Sensor electrode layout showing four top electrodes (red E1-E4) and one bottom one (blue). Electric fields couple directly between the top and bottom electrodes (X1), while some fringing fields (X2, X3) extend above the plane of the device and can couple into a finger for proximity detection. The device is a mutual capacitive sensor in which in **(D)** an applied pressure displaces the top electrodes (originally grey) downwards (red) to increase coupling with the bottom electrode (blue), while **(E)** shear is detected by the lateral displacement and varying overlap of the top and bottom electrodes. In **(F)** the structure of the sensor is shown. **(G)** Cross-section of sensor showing localized buckling upon shearing with a finger.

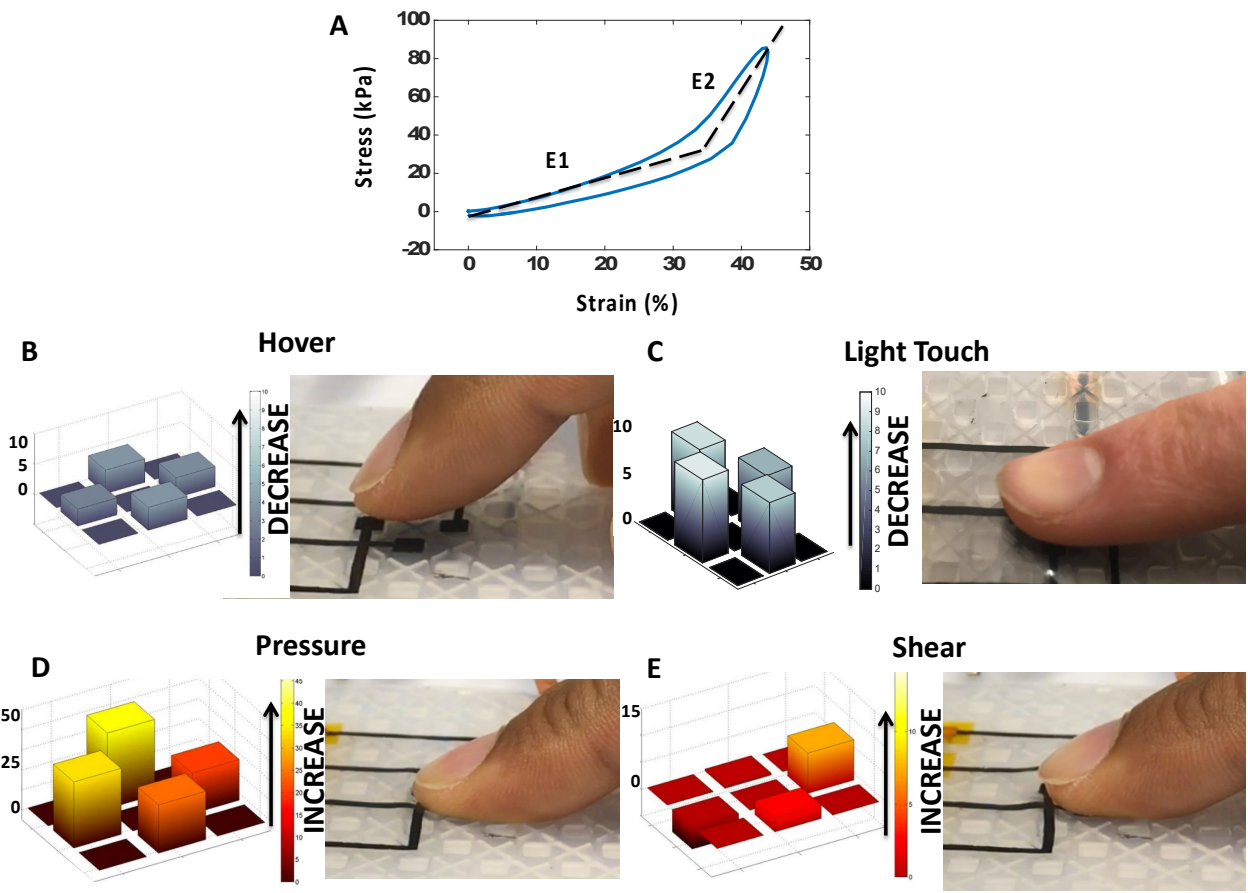

**Fig. 2. Sensor response to different stimuli. (A)** Non-linear elastic response similar to human skin. *E1* is a smaller effective elastic modulus for small strains while *E2* is a larger elastic modulus for larger strains. **(B)** Response to a hovering finger **(C)** a light touch **(D)** a press and **(E)** a shear. All capacitive axes are in %ΔC.

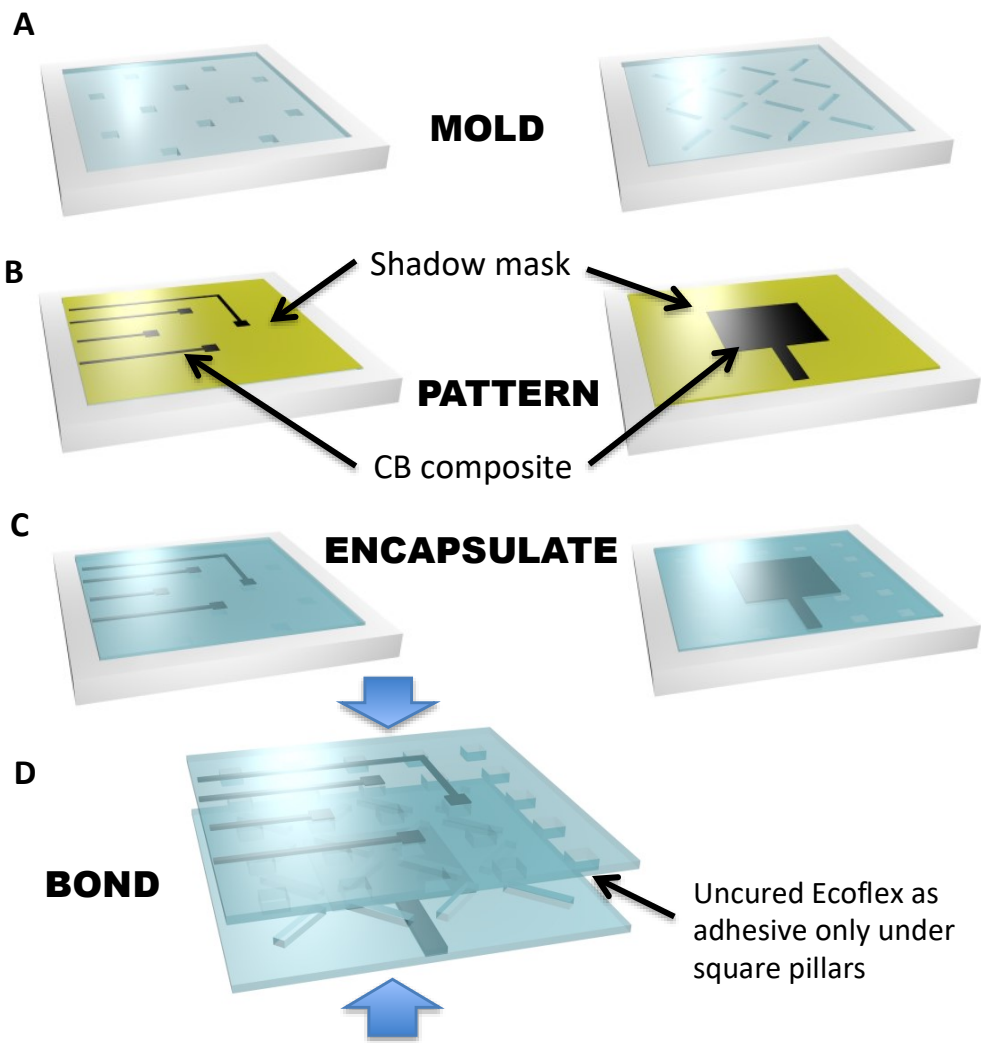

**Fig. 3. Fabrication of the sensor. (A)** Step 1: Mold. Mold filled with Ecoflex™, with the square pillars that will form the top surface of the skin shown at left and the X-shaped supports that form the bottom layer shown at right. **(B)** Step 2: Pattern electrodes (black) using carbon black- Ecoflex™ composite by doctor blading through a shadow mask. **(C)** Spin on encapsulating layer to prevent external electrical contact. **(D)** Step 3: Bond layers using a thin layer of uncured Ecoflex™ which glues the square pillars to the base layer.

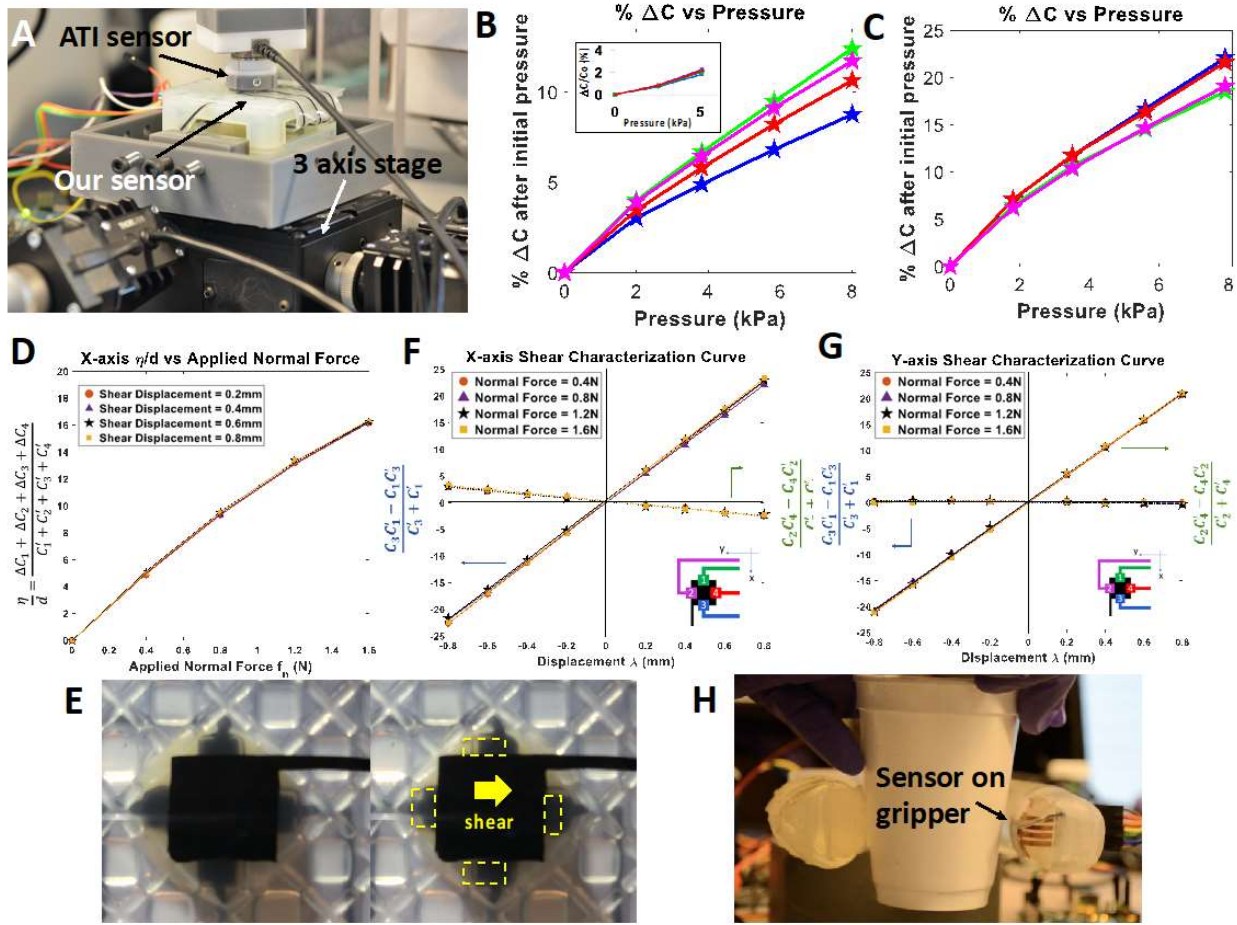

**Fig. 4. Characterization of sensor. (A)** Characterization setup showing the ATI Nano 17 load cell and the Thorlabs 3-axis stage. **(B)** Plot showing relation between applied pressure and %ΔC change in capacitance. The inset shows the simulated %ΔC change in capacitance with pressure. **(C)** Plot showing relation between applied pressure and % change in capacitance with GND shielding to reduce parasitic capacitance. **(D)** Plots showing response of sensor to normal forces with simultaneous horizontal forces being applied. **(E)** Image from the bottom of the sensor at steady state (left) and with a shear force applied (right). **(F)** Plots showing response of sensor to horizontal shear displacement in the X-axis with simultaneous normal forces being applied. **(G)** Plots showing response of sensor to horizontal shear displacement in the Y-axis with simultaneous normal forces being applied. **(H)** Sensor applied on a robotic gripper holding a paper cup.

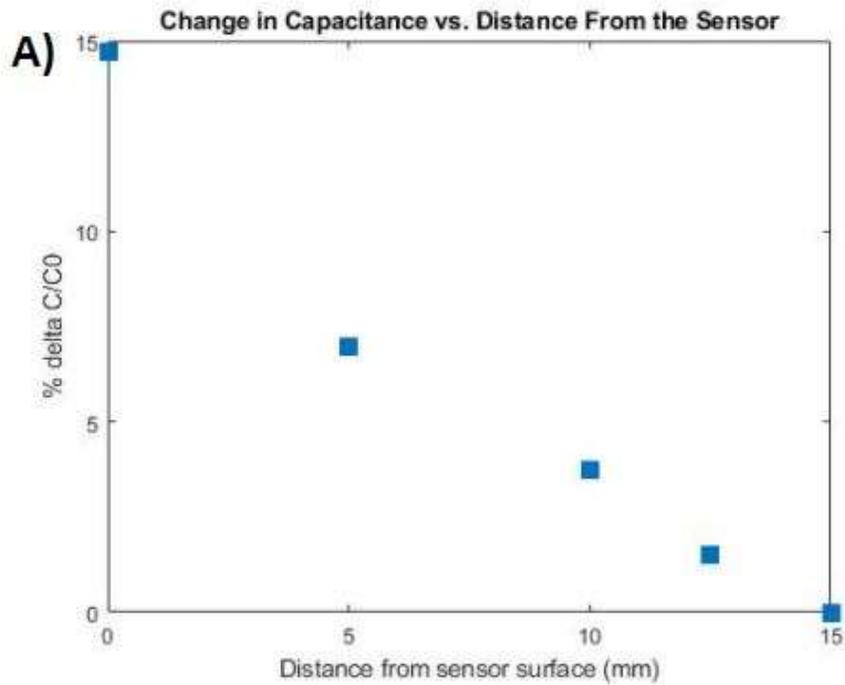
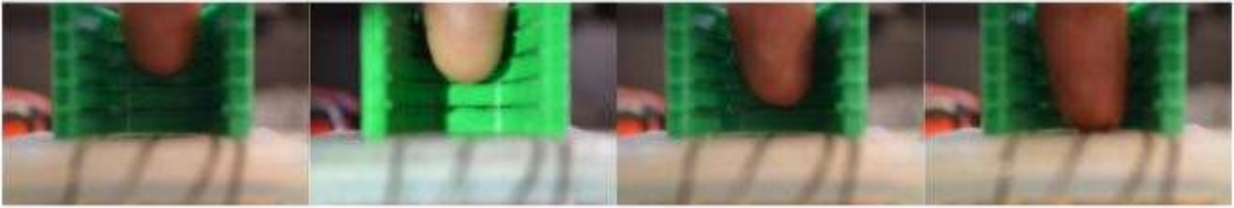

**Fig. 5. Approaching Object Proximity Characterization. (A)** Plot showing average taxel capacitive response to an approaching finger with 15 mm distance as baseline capacitance **(B)** Finger approaching sensor within a plastic distance-measuring guide

**Table 1. Truth table for proximity/light touch, two-axis shear and normal force.**

| Stimulus | C1 | C2 | C3 | C4 |
|---|---|---|---|---|
| Light touch | Down | Down | Down | Down |
| Pressure | Up | Up | Up | Up |
| Shear +ve X direction | Down | Unchanged | Up | Unchanged |
| Shear -ve X direction | Up | Unchanged | Down | Unchanged |
| Shear +ve Y direction | Unchanged | Up | Unchanged | Down |
| Shear -ve Y direction | Unchanged | Down | Unchanged | Up |

## SUPPLEMENTARY MATERIALS

Fig. S1. Calculating shear with pressure applied.
Fig. S2. Absolute change in capacitance VS pressure applied.
Fig. S3. $\Delta C/C_0$ VS pressure showing sensitivity.
Fig. S4. Pressure VS Strain for a range of displacements.
Fig. S5. Change in capacitance with strain.
Fig. S6. (A)Cross section of sensor showing parasitic coupling of interconnecting trace with top electrode (B) GND shield blocking parasitic coupling with interconnecting trace.
Fig. S7. (A) Flex PCB with bottom electrodes of sensor as copper pads on the PCB (B) top part of the sensor with stretchable dielectric and top carbon composite electrodes bonded on bottom electrodes on flex PCB (C) 3D model of sensor on a robot fingertip.
Fig. S8. Ambiguity due to proximity and small pressure.
Fig. S9. Conductor plane multiplexed to FDC1004 for self-capacitive proximity sensing.
Fig. S10. Sensor response to shear displacement and force.
Movie S1. Sensor demonstrating normal and horizontal force with angle being physically stimulated by a soft strawberry.
Movie S2. Sensor on a robotic gripper demonstrating detection of shear forces while pouring water in a Styrofoam cup.
Movie S3. Sensor with self-capacitive augmentation demonstrating multiple axes force detection and resolution of ambiguity between proximity and small pressure.

# scientific reports

## Supplementary Materials for

### Soft artificial skin for detecting approach, contact and shear

Mirza S. Sarwar[1]*, Ryusuke Ishizaki[2], Kieran Morton[1] Claire Preston[1], Tan Nguyen[1], Xu Fan[1], Bertille Dupont[1], Leanna Hogarth[1], Takahide Yoshiike[2], Shahriar Mirabbasi[1], John D.W. Madden[1]*

*Corresponding author. Email: mirzas@ece.ubc.ca, jmadden@ece.ubc.ca

**This PDF file includes:**
Supplementary Text
Figs. S1 to S17
Movies S1 to S3

**Other Supplementary Materials for this manuscript include the following:**
Movies S1 to S3

**Supplementary Text**

1      Analytical Equations to Differentiate Between Shear and Normal Pressure

In order to calculate the shear displacement and normal displacement from the sensor output, we derive analytical equations for shear displacement that are independent of the normal displacement and similarly, an equation for the normal displacement that is independent of the shear displacement. Fig. S1 displays the position of the sensor electrodes under a combined applied normal and shear force applied, and is used to guide these equations.

1.1      Separating shear from pressure

In deriving equations (1), (2) and (3) in main text, a parallel plate model is assumed. In this model, the undeformed state (Fig. S1 left) is characterized by a distance, $d$, between the top plates (red) and the bottom plate (blue), and an overlap area, $A$. The left and right capacitances ($C_1$ and $C_2$) are then equal, and estimated to be:

$$C_1 = \frac{\varepsilon A}{d}, C_2 = \frac{\varepsilon A}{d}. \quad (S1)$$

Here, the permittivity is $\varepsilon$. When a shear displacement, $\lambda$, and a simultaneous normal displacement, $\eta$, are applied, there is both a change in overlap area and distance between plates. Neglecting any fringing fields, the new capacitances are:

$$C_1' = \frac{\varepsilon(W \times (L+\lambda))}{d-\eta}, C_2' = \frac{\varepsilon(W \times (L-\lambda))}{d-\eta}. \quad (S2)$$

$W$ is the width of the electrode, into the page. The separation between plates following deformation is computed for the left and right capacitances:

$$(d-\eta) = \frac{\varepsilon(W \times (L+\lambda))}{C_1'} \quad (S3), \text{ and}$$

$$(d-\eta) = \frac{\varepsilon(W \times (L-\lambda))}{C_2'}. \quad (S4)$$

Setting (S3) and (S4) equal, the resulting equation is:

$$\frac{\varepsilon WL}{C_1'} + \frac{\varepsilon W\lambda}{C_1'} - \frac{\varepsilon WL}{C_2'} + \frac{\varepsilon W\lambda}{C_2'} = 0,$$

which is then multiplied by $1/d$ to obtain:

$$\frac{C_1}{C_1'} + \frac{\varepsilon W\lambda}{dC_1'} - \frac{C_2}{C_2'} + \frac{\varepsilon W\lambda}{dC_2'} = 0.$$

This is rearranged to give:

$$\frac{\varepsilon W\lambda}{d}\left(\frac{1}{C_1'} + \frac{1}{C_2'}\right) = \frac{C_2}{C_2'} - \frac{C_1}{C_1'}.$$

The resulting shear strain is then:

$$shear\ strain \rightarrow \frac{\lambda}{d} = \frac{C_2 C_1' - C_1 C_2'}{\varepsilon W(C_2' + C_1')}. \quad (S5)$$

This is Equation (2). Equation (3) is derived in the same way. The shear depends on the difference between capacitances, as expected.

## 1.2 Separating pressure from shear

In order to derive Equation (1), the sum of all four capacitances is used. Using a parallel plate capacitance once again, the four original capacitances are:

$$C_1 = \frac{\epsilon WL}{d}, C2 = \frac{\epsilon WL}{d}, C_3 = \frac{\epsilon WL}{d}, \text{and } C_4 = \frac{\epsilon WL}{d}. \quad \text{(Eqn. S6)}$$

The four capacitances in the deformed state are:

$$C_1' = \frac{\epsilon W(L+\lambda)}{d-\eta}, C_2' = \frac{\epsilon W(L-\lambda)}{d-\eta}, C_3' = \frac{\epsilon WL}{d-\eta}, \text{and } C_4' = \frac{\epsilon WL}{d-\eta}. \quad \text{(Eqn. S7)}$$

The sum of the initial and deformed capacitances are given by:

$$C_1 + C_2 + C_3 + C_4 = \frac{4\epsilon WL}{d} \text{ and,}$$
$$C_1' + C_2' + C_3' + C_4' = \frac{4\epsilon WL}{d-\eta}. \quad \text{(Eqn. S8)}$$

Taking the ratio of these two sums, and rearranging,

$$\frac{d-\eta}{d} = \frac{C_1+C_2+C_3+C_4}{C_1'+C_2'+C_3'+C_4'}, \text{ or}$$
$$\frac{\eta}{d} = 1 - \frac{C_1+C_2+C_3+C_4}{C_1'+C_2'+C_3'+C_4'}. \quad \text{(Eqn. S9)}$$

The strain in the dielectric is then:

$$\frac{\eta}{d} = \frac{C_1'-C_1+C_2'-C_2+C_3'-C_3+C_4'-C_4}{C_1'+C_2'+C_3'+C_4'}, \text{ or}$$
$$normal\ strain \to \frac{\eta}{d} = \frac{\Delta C_1 + \Delta C_2 + \Delta C_3 + \Delta C_4}{C_1'+C_2'+C_3'+C_4'}. \quad \text{(Eqn. S10)}$$

S10 is Equation 1 in the main text.

Not included in the equations are the effects of changing effective dielectric constant as the sensors are displaced, the effect of differences in dimensions between capacitors, and the effects of fringing. There is also an assumption that forces are uniform across the taxel. The relatively simple model is nevertheless effective in largely separating shear from normal force.

## 2 Change in Capacitance with Pressure

### 2.1 Absolute change in capacitance

Fig. S2 shows the magnitude of absolute change of the capacitance with pressure. It is observed that all four capacitors coincide relatively well. In Fig. 4B on the other hand, the four $\Delta C/C_0$ do responses are quite scattered. This is because, although the change in capacitance is similar for all four, the baseline values are different, and so the % change values are also different. This scattering is dramatically reduced by adding ground shielding, reducing the variation between base capacitances, as seen in Fig. 4C.

### 2.2 Sensitivity

In order to obtain the sensitivity over a region, a straight line is fit, as shown in Fig. S3. Based on the slope, the sensitivity at this level of stress is 1.5 %/kPa. The sensitivity is highest at small strains, and drops with increasing load.

2.3    Mechanical characterization

A series of stress-strain tests of varying amplitudes were applied to the sensor to observe its mechanical properties. A displacement-controlled sine wave of 0.1 Hz is applied. The stress vs. strain curves for different values of displacement amplitude are plotted in Fig. S4.

The apparent hysteresis between the loading and unloading parts of the cycle at large strain amplitudes suggests a viscoelastic behavior of the structure. The increase in stiffness at large amplitudes shows the non-linear response of the structure. The effective elastic modulus is higher for larger strains. A similar non-linear behavior in elastic modulus with strain is observed in human skin, shown by Delalleau *et al.* [21].

Plotting $\Delta C/C_0$ against applied strain for a range of different displacements is shown in Fig. S5. There is very little apparent hysteresis even at large strains, showing that the capacitance is largely determined by displacement. This is expected, since capacitance is a function of geometry.

3    Ground shield to reduce parasitic coupling with interconnecting trace

There is capacitive coupling between electrical connections and the sense electrodes, as depicted in Fig. S6A. The coupling also occurs between interconnects, as discussed in part 3 above. A ground shield architecture is shown in Fig. S6B that reduces this coupling. The resulting relative change in capacitance then becomes more consistent between electrodes, as shown by comparing Fig. 4B (unshielded) and Fig. 4C (shielded).

4    Flex-Stretch sensor implementation

In order to enable higher electrode resolution and relative alignment, a flex printed circuit board can act as a base for the sensor, as shown in Fig. S7. The electrodes are now a part of the readout hardware, also providing a shielded path for the connections when a multilayer board is used.

5    Ambiguity due to proximity and small pressure

Combining both projected (mutual capacitance) and overlap capacitance in the same sensor leads to a small region of ambiguity, as explained in Fig. S8. In Case A, when a finger comes in close to a sensor (proximity) there is a small decrease in capacitance. The magnitude of the capacitance reaches its minimum value when a light contact is made. Retraction of the finger leads it to rise again. If the finger is still nearby, the capacitance will still be lower than the baseline value. This lowering can be interpreted as finger proximity. In Case B, when the finger presses down slightly, there is a small increase in capacitance due to the decrease in the dielectric thickness. The capacitance change looks identical to a proximity response. There are cases when proximity and a light press cannot be distinguished.

In order to resolve this ambiguity, a ground plane can be used to shield the effects of the presence of a finger near the sensor. Although this solves the ambiguity issue, the sensor is now only able to detect pressure and shear. The proximity sensing ability is removed.

In order to reintroduce the proximity sensing ability, we now multiplex the top layer between a ground connection (when reading the pressure/shear response) and a self-capacitance readout IC (e.g. FDC1004), which can detect the proximity/light contact of a finger. A diagram of this architecture is shown in Fig. S9, with the upper ground plane preventing outside fields from causing sensor ambiguity.

6       Sensor response of individual electrode capacitance to shear

Fig. S10 shows the response of individual electrode capacitances of the sensor when a horizontal displacement or force is applied with a simultaneous normal force. It is observed that all four plots of %$\Delta C/C_0$ (under various normal forces) coincide with each other. This means that the shear response is largely independent of normal force applied. The x-axis shear response is also decoupled from shear in the y direction. In case of %$\Delta C/C_0$ with shear force, we see a small deviation in the case of the smallest normal force applied (i.e. 0.4N).

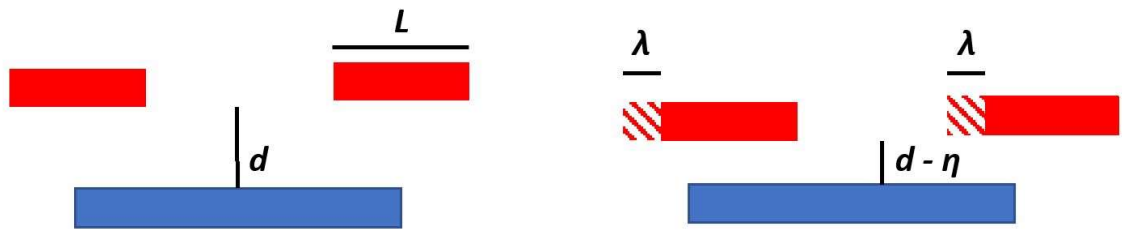

**Fig. S1.** Calculating shear with pressure applied.

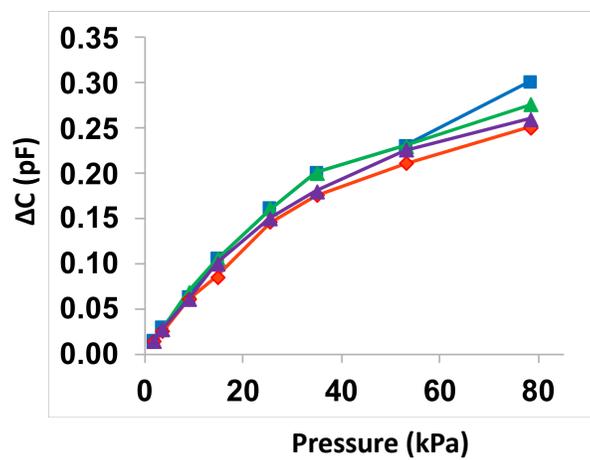

**Fig. S2. Absolute change in capacitance VS pressure applied.**

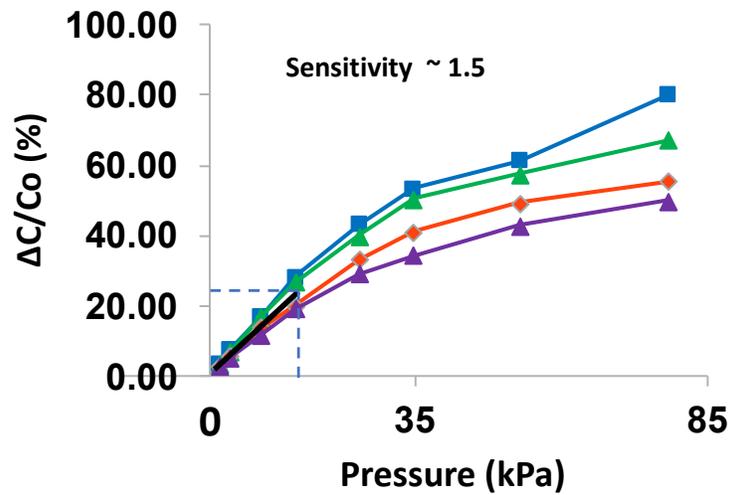

**Fig. S3.** ΔC/C$_0$ *vs*. pressure showing sensitivity of about 1.5%/kPa in this case.

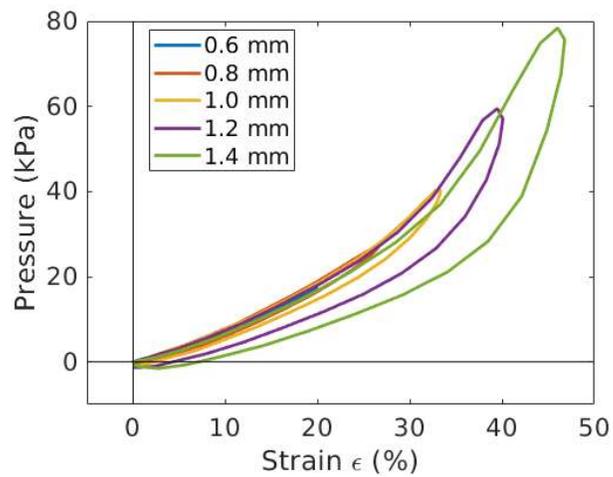
**Fig. S4. Pressure vs. strain for a range of displacements**

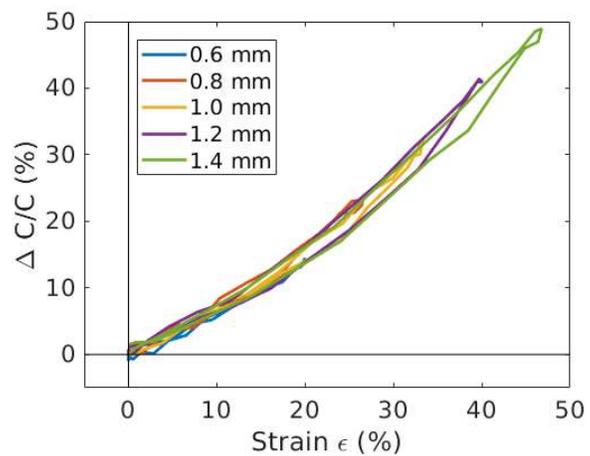

**Fig. S5.** Measured change in capacitance with strain.

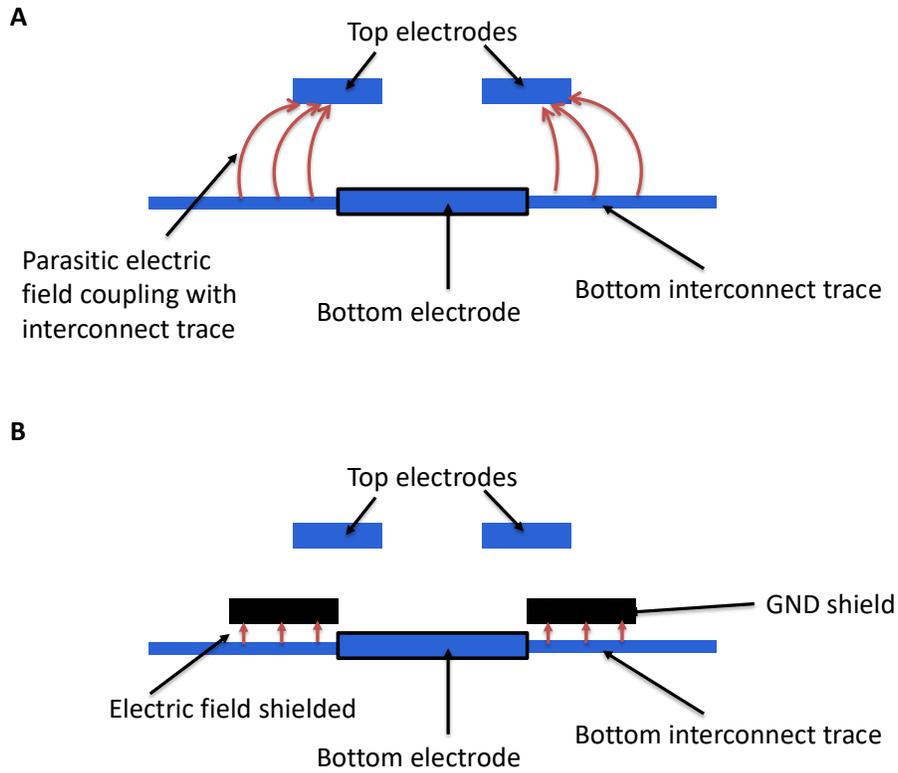

**Fig. 6.** (A) Cross-section of sensor showing parasitic coupling of interconnected traces (left and right bottom) with top electrode. There is also coupling between top and bottom connection traces. (B) A ground shield is introduced to block parasitic coupling with interconnected trace, reducing the overall capacitance of the sensor, and increase the relative change in capacitance.

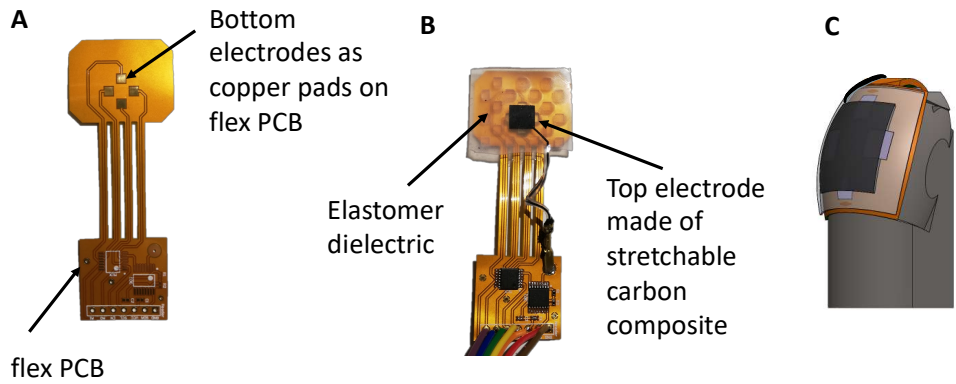

**Fig. S7.** (A) Flex-PCB with bottom electrodes of the sensor as copper pads on the PCB, and (B) Top part of the sensor with stretchable dielectric and top carbon composite electrodes bonded on bottom electrodes on flex-PCB (C) 3D model of sensor on a robot fingertip.

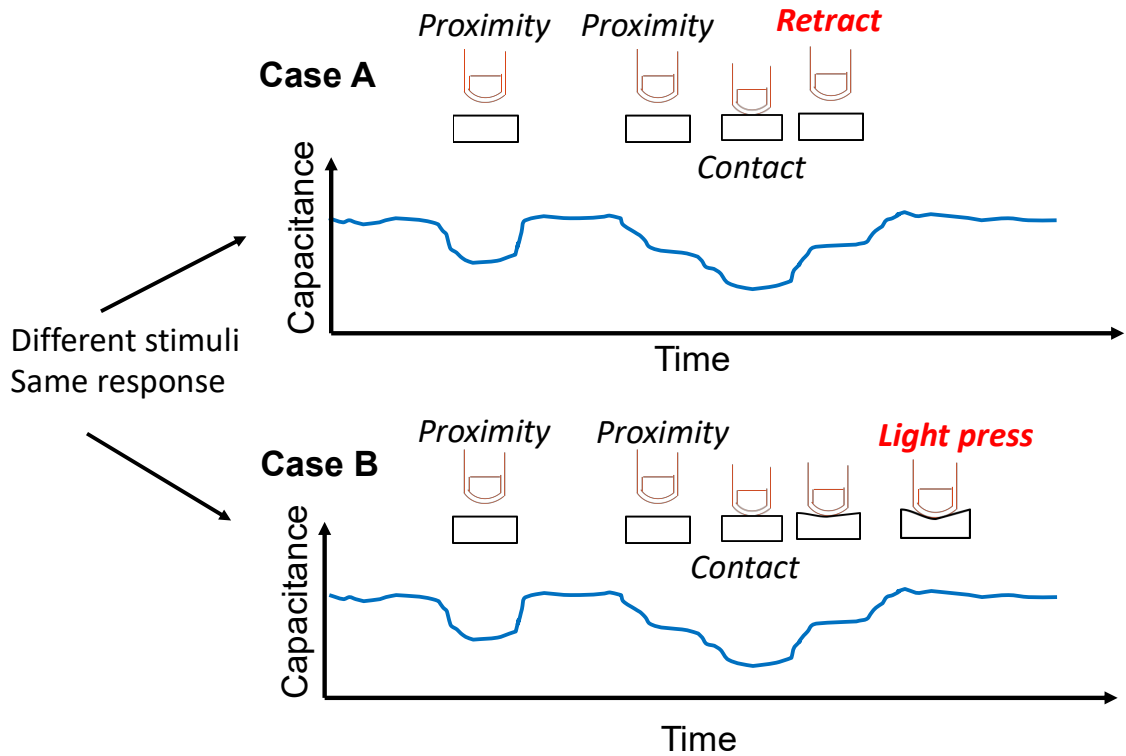

**Fig. S8.** Ambiguity of capacitive response between proximity and light contact. Case A (top): proximity leads to a decrease in capacitance, which becomes smallest upon light contact. The capacitance rises again upon partial retraction. Case B (bottom): contact followed by the application of force, which brings the capacitor plates closer, leads to increase in capacitance. The two steps in red look the same to the sensor, creating an ambiguity.

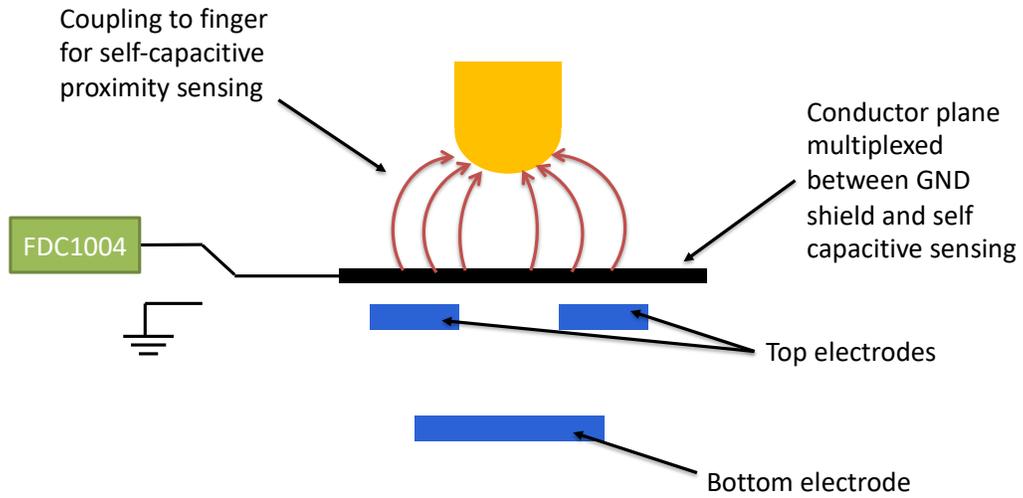

**Fig. S9.** Conductor plane multiplexed to FDC1004 for self-capacitive proximity sensing. The ambiguity presented in Fig. S8 can be resolved by shielding the sensor from proximity, and separately detecting proximity using the top electrode.

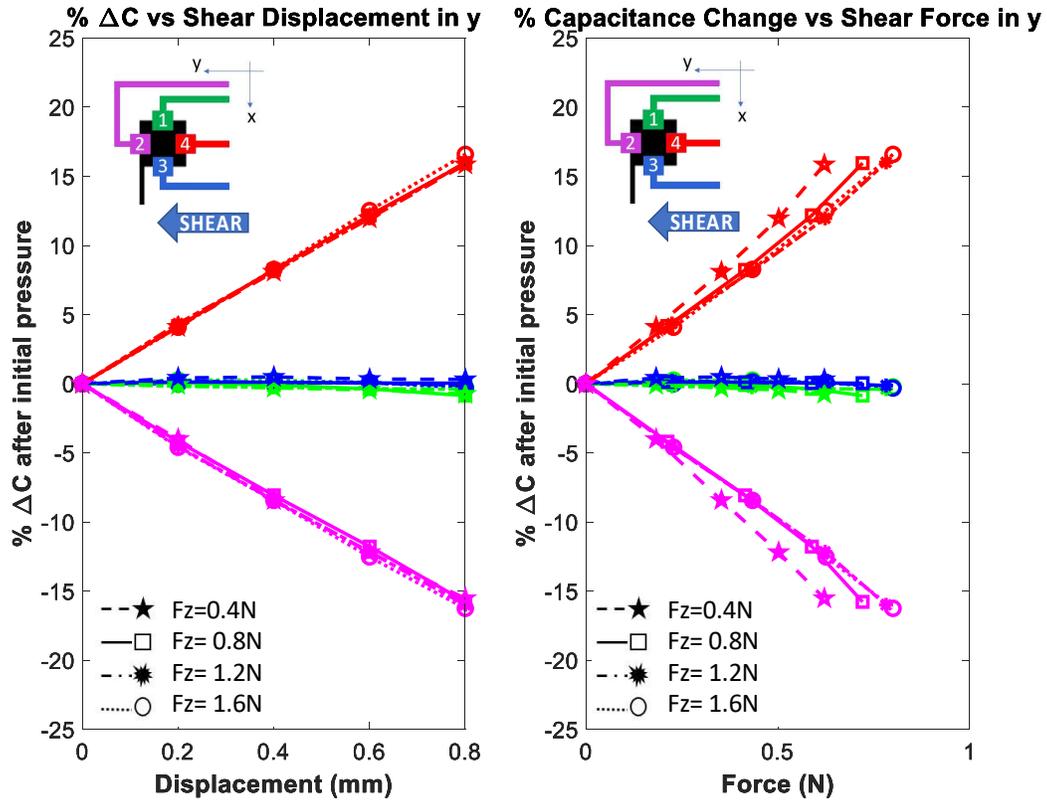

**Fig. S10.** Sensor response to shear displacement and force. The plots show that sensitivity to shear displacement (left) and shear force (right) are essentially unaffected by normal force. The x-axis shear is decoupled from the y-axis. The forces in the legend represent the constant normal forces applied during shear tests. The horizontal axes are shear displacement (left) and shear force (right). The vertical axis is relative change in capacitance.

**Movie S1.** Measurement of pressure and shear using the sensor, with force applied by a soft strawberry.

**Movie S2.** Change in pressure and shear applied by a cup to a robotic finger as a result of filling the cup with water.

**Movie S3.** Sensor differentiation of proximity, touch, pressure, and shear.